\documentclass{article}
\usepackage{proceedings}

\usepackage{amssymb}
\usepackage{latexsym}
\usepackage{amsmath}
\usepackage[all]{xy}
\newcommand{\NewTh}[2]{ %
	\newtheorem{#2}{#2} %
	\newenvironment{#1}{\begin{#2} }{~\hfill$\Box$\end{#2}} %
}
\NewTh{df}{Definition}
\NewTh{ex}{Example}
\NewTh{Th}{Theorem}
\NewTh{co}{Corollary}
\NewTh{lem}{Lemma}
\NewTh{ppy}{Property}
\NewTh{ppn}{Proposition}

\def\eqdef{\stackrel{\rm def}{=} \;}
\def\L3{L_3}
\def\Not{\hbox{\rm \em not} \;}
\def\l{\hbox{\rm \bf \em l} \,}
\def\m{\hbox{\rm \bf \em m} \,}
\def\x{\circ \,}
\def\y{\bullet \,}
\def\S{\Sigma}
\def\u{1\!/\!2}
\def\true{\top}
\def\false{\bot}
\def\eq{\leftrightarrow}
\def\EQ{\Leftrightarrow}

\newcommand{\PAR}[1]{ \big( #1 \big) }
\newcommand{\ta}[1]{\dot{#1}}
\newcommand{\tb}[1]{\ddot{#1}}
\def\CIRC{\hbox{\rm CIRC}}
\def\SUBT{\hbox{\rm SUBT}}

\newcommand{\APPEND}[2]{%
   \section*{Appendix #1. #2.}%
   \markboth{Appendix #1}{#2}%
   \renewcommand{\thesection}{#1-\arabic{section}}%
   \setcounter{section}{0}%
   \renewcommand{\theequation}{#1-\arabic{equation}}%
   \setcounter{equation}{0}%
}

\title{Alternative Characterizations for \\ Strong Equivalence of Logic Programs}

\author{Pedro Cabalar\\
    \\
    AI Lab., Dept. of Computer Science\\
    University of Corunna, Spain\\
    {\tt cabalar@dc.fi.udc.es}
}

\begin{document}

\maketitle

%%%%%%%%%%%%%%%%%%%%%%%%%%%%%%%%%%%%%%%%%%%%%%%%%%%%%%%%%%%%%%%%%%%%%%%%%%%%%%%
\begin{abstract}
In this work we present additional results related to the property of strong equivalence of logic programs. This property asserts that two programs share the same set of stable models, even under the addition of new rules. As shown in a recent work by Lifschitz, Pearce and Valverde, strong equivalence can be simply reduced to equivalence in the logic of {\em Here-and-There} (HT). In this paper we provide two alternatives respectively based on classical logic and 3-valued logic. The former is applicable to general rules, but not for nested expressions, whereas the latter is applicable for nested expressions but, when moving to an unrestricted syntax, it generally yields different results from HT.
\end{abstract}

%%%%%%%%%%%%%%%%%%%%%%%%%%%%%%%%%%%%%%%%%%%%%%%%%%%%%%%%%%%%%%%%%%%%%%%%%%%%%%%
\section{Introduction}
%%%%%%%%%%%%%%%%%%%%%%%%%%%%%%%%%%%%%%%%%%%%%%%%%%%%%%%%%%%%%%%%%%%%%%%%%%%%%%%

There is no doubt that the application of logic programming (LP) as a tool for knowledge representation has influenced in the progressive evolution of LP towards a more logical-style orientation, avoiding the initial syntactic restrictions. Think, for instance, how the stable models semantics~\cite{Gel88} has been successively modified to cope with explicit negation and disjunctive heads~\cite{Gel91}, default negation in the head~\cite{Lif94,Ino98} or, finally, the full use of nested expressions~\cite{Lif99}. Perhaps as a result of this evolution, the following question has become interesting: when can we consider that two (syntactically different) programs $\Pi_1$ and $\Pi_2$ {\em represent the same knowledge}?

From a traditional LP perspective, we would say that $\Pi_1$ and $\Pi_2$ are equivalent when they share the same set of stable models like, for instance, the programs $\{p\}$ and $\{p \leftarrow \Not q\}$ whose only stable model is $\{p\}$. However, nonmonotonicity may cause them to behave in a different way in the presence of additional rules (just add fact $q$ to both programs). Thus, if we want to check whether $\Pi_1$ and $\Pi_2$ actually represent the {\em same knowledge}, we must require a stronger condition, talking instead about {\em strong equivalence}: for any $\Pi$, the stable models of $\Pi_1 \cup \Pi$ and $\Pi_2 \cup \Pi$ coincide.

An elegant solution to this problem is the recent characterization of stable models relying on Heyting's logic of {\em Here-and-There} (HT). In~\cite{Pea97}, Pearce first showed that stable models can be simply seen as some kind of minimal HT models. Then, in~\cite{Lif00}, Lifschitz, Pearce and Valverde proved that, in fact, this characterization fits with the semantics for nested operators independently proposed in~\cite{Lif99} and, what is more important, that two programs are strongly equivalent iff they have the same set of HT models.

In this paper we provide two closely related alternatives to HT that rely on classical logic and 3-valued logic ($\L3$), respectively. These alternatives present the advantage of using very well-known formalisms, which may help for a better insight of strong equivalence (the main emphasis of this paper), but can be useful for implementation purposes too. Unfortunately, we also show how, in both cases, their scope of applicability seems to be smaller than in the HT case. This is evident for the classical encoding we present, which can only be used as a ``direct" semantics\footnote{Application of the classical encoding to nested expressions is also possible, but only after a previous {\em syntactic} transformation.} for non-nested logic programs, whereas the $\L3$ characterization properly handles nested expressions in a direct way, but loses some important properties when nesting is also allowed for rule conditionals.

The paper is structured as follows. The next section recalls the basic definition of stable models for general (non-nested) logic programs. Section \ref{sec:class} describes the classical encoding. In Sections \ref{sec:nest} and \ref{sec:3v} we respectively describe nested expressions and the 3-valued formalization. After that, we briefly comment the differences between the HT and $\L3$ interpretations. Finally, Section \ref{sec:conc} concludes the paper. Proofs of theorems have been included in an appendix.

%%%%%%%%%%%%%%%%%%%%%%%%%%%%%%%%%%%%%%%%%%%%%%%%%%%%%%%%%%%%%%%%%%%%%%%%%%%%%%%
\section{Stable models}
%%%%%%%%%%%%%%%%%%%%%%%%%%%%%%%%%%%%%%%%%%%%%%%%%%%%%%%%%%%%%%%%%%%%%%%%%%%%%%%
\label{sec:sm}
The syntax of logic programs is defined starting from a finite set of ground atoms $\S$, called the {\em Herbrand base}, which will serve as propositional signature. We assume that all the variables have been previously replaced by their possible ground instances. Letters $a, b, c, d, p, q$ will be used to denote atoms in $\S$, and letters $I, J$ to denote subsets of $\S$. A {\em logic program} is defined as a collection of rules of the shape:
\begin{multline}
a_1 ; \dots ; a_m ; \Not b_1 ; \dots ; \Not b_n \leftarrow \\
c_1 , \dots , c_r , \Not d_1 , \dots , \Not d_s \label{f:lprule}
\end{multline}
We call {\em head} (resp. {\em body}) to the left (resp. right) hand side of the arrow in (\ref{f:lprule}). The comma and the semicolon are alternative representations of conjunction $\wedge$ and disjunction $\vee$, respectively. When $m=n=0$ we usually write $\bot \leftarrow B$ instead of $\leftarrow B$, whereas when $r=s=0$ we directly write $H$ instead of $H \leftarrow$ or $H \leftarrow \top$.

Sometimes, it will be convenient to think about program rules as classical propositional formulas, where $\leftarrow$ and $\Not$ are respectively understood as material implication and classical negation. In this way, the usual expression $I \models R$ denotes that interpretation $I$ satisfies rule $R$ (seen as a classical formula), whereas $I \models \Pi$ means that $I$ is a model of the program $\Pi$ (seen as a classical theory).

The {\em reduct} of a program $\Pi$ w.r.t. some set of atoms $I$, written $\Pi^I$, is defined as the result of replacing in $\Pi$ any default literal $\Not p$ by $\top$, if $p \not\in I$, or by $\bot$ otherwise.

\begin{df}{\bf (Stable model)}
A set of atoms $I \subseteq \S$ is a {\em stable model} of a logic program $\Pi$ iff $I$ is a minimal model of $\Pi^I$.
\end{df}

%%%%%%%%%%%%%%%%%%%%%%%%%%%%%%%%%%%%%%%%%%%%%%%%%%%%%%%%%%%%%%%%%%%%%%%%%%%%%%%
\section{Strong equivalence in classical logic}
%%%%%%%%%%%%%%%%%%%%%%%%%%%%%%%%%%%%%%%%%%%%%%%%%%%%%%%%%%%%%%%%%%%%%%%%%%%%%%%
\label{sec:class}

We can think about the definition of stable models as a try-and-error procedure which handles (propositional) interpretations for two different purposes. On the one hand, we start from some arbitrary interpretation $I^a$, we can call the initial {\em assumption}, used to compute the reduct $\Pi^{I^a}$. On the other hand, in a second step, we deal with minimal models of $\Pi^{I^a}$ which, in principle, {\em need not to have any connection} with $I^a$. Each minimal model  $I^p$ can be seen as the set of propositions we can {\em prove} by deductive closure using the rules in $\Pi^{I^a}$. When the proved atoms coincide with the initial assumption, $I^p = I^a$, a stable model is obtained.

In order to capture this behavior, we reify all the atoms $p \in \S$ to become arguments of two unary predicates, $assumed(p)$ and $proved(p)$, that respectively talk about $I^a$ and $I^p$. Sort variable $X$ will be used for ranging over any propositional symbol in $\S$. When considering the models of any reified formula $F$, we will implictly assume that they actually correspond to $F \wedge \hbox{UNA}$, where UNA stands for the unique names assumption for sort $\S$. This allows us identifying any Herbrand model $M$ of this type of formulas with a pair\footnote{The superscripts $p$ and $a$, which stand here for {\em proved} and {\em assumed}, respectively correspond to the worlds {\em here} and {\em there} in HT or to the sets of {\em positive} and {\em non-negative} atoms in $\L3$.} $(I^p,I^a)$ so that $M[assumed]=I^a$ and $M[proved]=I^p$. Expression $M \models F$ represents again satisfaction of reified formulas -- ambiguity with respect to $I \models F$ is cleared by the shape of structures and formulas.

Given this simple framework, we provide two encodings: the first one is a {\em completely straightforward} translation to capture stable models, whereas the second one is a stronger translation to characterize strong equivalence.

\begin{df}{\bf (First translation)}
\label{df:1}
For any logic program rule $R$ like (\ref{f:lprule}), we define the classical formula $\ta{R}$ as the material implication:
\begin{multline}
\PAR{\bigwedge^r_{i=1} proved(c_i)} \wedge \PAR{\bigwedge^s_{i=1} \neg assumed(d_i)} \supset \\  \PAR{\bigvee^n_{i=1} proved(a_i) } \vee \PAR{\bigvee^m_{i=1} \neg assumed(b_i)} \label{f:circ-1}
\end{multline}
Given a logic program $\Pi$, the formula $\ta{\Pi}$ stands for the conjunction of all the $\ta{R}$, for each rule $R \in \Pi$. 
\end{df}

Intuitively, to obtain the minimal models $I^p$ of $\Pi^{I^a}$ we can use an ordering relation among pairs $(I^p,I^a) \preceq (J^p,J^a)$ that holds when both $I^a=J^a$ is fixed and $I^p \subseteq J^p$. The corresponding models $\preceq$-minimization have a simple syntactic counterpart\footnote{See Section 2.5 in~\cite{Lif93}.}: predicate circumscription $\CIRC[\ta{\Pi};proved]$. Second, after obtaining minimal models, we must further require $I^p=I^a$, that is, we want pairs of shape $(I,I)$ where what we assumed results to be exactly what we proved. These pairs of shape $(I,I)$ will be called {\em total}. Clearly, forcing models to be total corresponds to including of the formula:
\begin{eqnarray}
\forall X. \ \PAR{proved(X) \equiv assumed(X)} \label{f:circ-2}
\end{eqnarray}

The intuitions above are not new. In fact, they were used in Theorem 5.2 in~\cite{Lin92} which, adapted\footnote{In~\cite{Lin92} they used a duplicated signature (atoms $p$ and $p'$) instead of reification and, therefore, they actually applied parallel circumscription. This result seems to have been first presented in~\cite{Lin91}.} to our current presentation, states the following result:

\begin{ppn}
Let $\S$ be a propositional signature. A set of atoms $I \subseteq \S$ is a stable model of a logic program $\Pi$ iff $M=(I,I)$ satisfies the formula:
\begin{eqnarray*}
\CIRC[\ta{\Pi};proved] \wedge (\ref{f:circ-2})
\end{eqnarray*}
\end{ppn}

In order to capture strong equivalence of two programs, it seems that we should not only compare the final selected models, but also the set of non-minimal ones involved in the minimization. For instance, it is easy to see that, due to monotonicity of classical logic, the following proposition trivially applies:

\begin{ppn}
\label{ppn:suff}
Let $\Pi_1$ and $\Pi_2$ be two logic programs such that $\models \ta{\Pi}_1 \equiv \ta{\Pi}_2$. Then $\Pi_1$ and $\Pi_2$ are strongly equivalent.
\end{ppn}

Unfortunately, the opposite does not necessarily hold: $\Pi_1$ and $\Pi_2$ can be strongly equivalent while $\ta{\Pi}_1$ and $\ta{\Pi}_2$ have different models. This is because encoding in Definition~\ref{df:1} allows some models which are actually irrelevant for strong equivalence, as we will show next.

\begin{df}{\bf (Subtotal model)}
For any reified theory $T$, a model $(I^p,I^a)$ of $T$, with $I^p \subseteq I^a$, is called {\em subtotal} iff $(I^a,I^a)$ is also model of $T$.
\end{df}

Let $\SUBT(T)$ represent the set of subtotal models of $T$ (note that total models are also included). It is clear that any model $M \not\in \SUBT(\ta{\Pi})$ is irrelevant for selecting the total $\preceq$-minimal models, i.e., for obtaining the stable models of $\Pi$. The next theorem shows that the coincidence of subtotal models is a necessary condition for strong equivalence. The proof (in the Appendix) constitutes a direct rephrasing of that for the main theorem in~\cite{Lif00}.
% and can be found together with the rest of proofs of this paper in the extended version~\cite{Cab01}.

\begin{Th}
\label{th:strong}
Two logic programs $\Pi_1$ and $\Pi_2$ are strongly equivalent iff $\ \ \ \SUBT(\ta{\Pi_1})=\SUBT(\ta{\Pi_2})$
\end{Th}

Theorem \ref{th:strong} points out that the $\ta{\Pi}$ encoding is still too weak for a full characterization of strong equivalence. We show next how, using a more restrictive translation (that is, adding more formulas) it is possible to obtain theories for which all their models are subtotal. To understand how to do this, consider the example program $\Pi_0=\{p \leftarrow q\}$ where:
\begin{eqnarray*}
\ta{\Pi}_0 & \eqdef & proved(q) \supset proved(p)
\end{eqnarray*}

This formula has 12 models: it restricts the extent of $proved$ to 3 cases ($\emptyset$, $\{p\}$ and $\{p,q\}$) leaving free, in each case, the 4 possibilities for $assumed$.

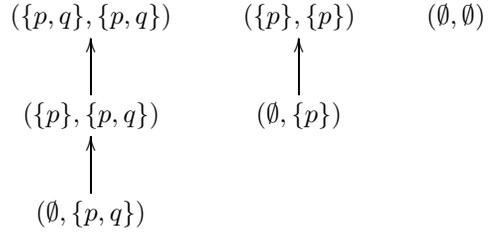
\begin{figure}[htbp]
\[
\xymatrix @-1mm {
   {(\{p,q\},\{p,q\})}
 & {(\{p\},\{p\})}
 & {(\emptyset,\emptyset)}
\\
   {(\{p\},\{p,q\})}
     \ar[u]_{}     
 & {(\emptyset,\{p\})}
     \ar[u]_{}     
 & *+\txt{}
\\ 
   {(\emptyset,\{p,q\})}
     \ar[u]_{}     
}
\]
\caption{Subtotal models of $\ta{\Pi}_0$.}
\label{fig:1}
\end{figure}

Figure~\ref{fig:1} shows the 6 subtotal models of $\ta{\Pi}_0$, representing the $\preceq$-ordering relationships among them. Notice how subtotal models always satisfy $I^p \subseteq I^a$, that is, we can require:
\begin{eqnarray}
\forall X. \PAR{proved(X) \supset assumed(X)} \label{f:circ-4}
\end{eqnarray}

Unfortunatley, the addition of this axiom is still not enough to rule out all the irrelevant models. For instance, $\ta{\Pi}_0 \wedge (\ref{f:circ-4})$ has still one non-subtotal model: $(\emptyset,\{q\})$. This model, however, has the particularity that its assumed atoms $I^a=\{q\}$ do not satisfy the original program rule: $I^a \not\models p \leftarrow q$. As it is well-known, any stable model $I$ of a program $\Pi$, is also a classical model: $I \models \Pi$. So, instead of starting from any arbitrary initial assumption $I^a$, we can begin requiring $I^a \models \Pi$. This can be easily incorporated into the encoding as follows. For each logic program rule $R$ like (\ref{f:lprule}), we define $\tb{R}$ as:
\begin{multline}
\PAR{\bigwedge^r_{i=1} assumed(c_i)} \wedge \PAR{\bigwedge^s_{i=1} \neg assumed(d_i)} \supset \\ \PAR{\bigvee^n_{i=1} assumed(a_i) } \vee \PAR{\bigvee^m_{i=1} \neg assumed(b_i)} \label{f:circ-3}
\end{multline}
Again, $\tb{\Pi}$ stands for the conjunction of $\tb{R}$ for all $R \in \Pi$.

\begin{df}{\bf (Second translation)}
For any logic program $\Pi$ we define the formula $\Pi^* \eqdef \ta{\Pi} \wedge \tb{\Pi} \wedge (\ref{f:circ-4})$.
\end{df}

The proof for the following theorem uses well-known properties of circumscription (see~\cite{Lif93}) to show that the additional formulas {\em do not affect} to the final set of stable models.

\begin{Th}
\label{th:circ}
For any logic program $\Pi$:\\
$\CIRC[\Pi^*;proved] \wedge (\ref{f:circ-2}) \equiv \CIRC[\ta{\Pi};proved] \wedge (\ref{f:circ-2})$
\end{Th}

But, of course, the real interest of $\Pi^*$ is that it finally rules out irrelevant models:
\begin{ppy}
\label{ppy:total}
Let $\Pi$ be a logic program. Then, any model $M=(I^p,I^a)$ of $\Pi^*$ is subtotal.
\end{ppy}

Finally, this property, together with theorem \ref{th:strong}, directly implies:

\begin{Th}
\label{th:strong2}
Two logic programs $\Pi_1$ and $\Pi_2$ are strongly equivalent iff $\ \ \ \models \Pi^*_1 \equiv \Pi^*_2$.
\end{Th}

%%%%%%%%%%%%%%%%%%%%%%%%%%%%%%%%%%%%%%%%%%%%%%%%%%%%%%%%%%%%%%%%%%%%%%%%%%%%%%%
\section{Nested expressions.}
%%%%%%%%%%%%%%%%%%%%%%%%%%%%%%%%%%%%%%%%%%%%%%%%%%%%%%%%%%%%%%%%%%%%%%%%%%%%%%%
\label{sec:nest}

The previous section has shown a way of reducing strong equivalence of logic programs into a simple equivalence test in classical logic, providing in this way a (we think) easier alternative to the HT characterization. However, although we gain in simplicity, it must be noticed that we lose in semantic quality: this classical encoding does not provide a general interpretation for program connectives but, instead, is limited to rules of shape (\ref{f:lprule}). The HT characterization is clearly stronger in this sense, since it provides a direct interpretation for {\em any} possible nesting of program connectives.

In~\cite{Lif99} a more general shape for program rules was considered. A {\em nested expression} is defined as any propositional combination of atoms with 0-ary operators $\bot$, $\top$, unary operator $\Not$ and binary operators `,' and `;'. A logic program is now a set of rules $Head \leftarrow Body$ where $Head$ and $Body$ are nested expressions (notice that the rule conditional $\leftarrow$ is the only operator that cannot be nested). An example of such a rule could be, for instance:
\begin{eqnarray}
a , b \leftarrow \Not (c ; \Not d) \label{f:nest-1}
\end{eqnarray}

Stable models for this kind of programs can be easily described by a simple modification in the definition of program reduct. We define now $\Pi^I$ as the result of replacing in $\Pi$ every maximal occurrence\footnote{That is, any $\Not F$ that is not in the scope of an outer $\Not$.} of $\Not F$ by $\bot$ if $I \models F$ or by $\top$ otherwise. Note that the previous definition of reduct corresponds to the particular case in which $F$ is an atom.

An interesting result derived from this modified semantics (proposition 7 in~\cite{Lif99}) is that any program with nested expressions is strongly equivalent to some (non-nested) program, just consisting of rules like (\ref{f:lprule}). To obtain this non-nested program, the following transformations are defined. Let $F, G$ and $H$ represent nested expressions. By $\alpha \EQ \beta$ we mean that we replace some regular occurrence of $\alpha$ by $\beta$. Then, we handle the following strongly equivalent transformations:
\begin{itemize}
\item[(i)] $F, G \EQ G, F$ and $F; G \EQ G; F$.
\item[(ii)] $(F, G), H \EQ F, (G, H)$ and \\ $(F; G); H \EQ F; (G; H)$.
\item[(iii)] $F, (G; H) \EQ (F, G); (F, H)$ and \\ $F; (G, H) \EQ (F; G), (F; H)$.
\item[(iv)] $\Not (F; G) \EQ \Not F, \Not G$ and \\ $\Not (F, G) \EQ \Not F; \Not G$.
\item[(v)] $\Not \Not \Not F \EQ \Not F$.
\item[(vi)] $F, \top \EQ F$ and $F; \top \EQ \top$.
\item[(vii)] $F, \bot \EQ \bot$ and $F; \bot \EQ F$.
\item[(viii)] $\Not \top \EQ \bot$ and $\Not \bot \EQ \top$.
\item[(ix)] $(F, G \leftarrow H) \EQ (F \leftarrow H), (G \leftarrow H)$.
\item[(x)] $(F \leftarrow G; H) \EQ (F \leftarrow G), (F \leftarrow H)$.
\item[(xi)] $(F \leftarrow G, \Not \Not H) \EQ (F; \Not H \leftarrow G)$.
\item[(xii)] $(F; \Not \Not G \leftarrow H) \EQ (F \leftarrow \Not G, H)$.
\end{itemize}

For instance, rule (\ref{f:nest-1}) can be successively transformed as follows:

\vspace{15pt}
\begin{tabular}{ll}
$a , b \leftarrow \Not c , \Not \Not d.$ & By (iv) \\
& \\
$a \leftarrow \Not c , \Not \Not d,$ & \\
$b \leftarrow \Not c , \Not \Not d.$ & By (ix) \\
& \\
$a ; \Not d \leftarrow \Not c,$ & \\
$b ; \Not d \leftarrow \Not c.$ & By (xi)
\end{tabular}
\vspace{15pt}

This treatment of nested expressions exceeds the applicability of our previous classical logic representation. From a practical point of view, such a limitation is not very important, since we can always unfold nested expressions by applying (i)-(xii). Nevertheless, from a theoretical point of view, this clearly points out that the classical encoding fails as a real semantic characterization for LP connectives.

As shown in~\cite{Lif00}, one of the important features of the HT formalization, apart from the result for strong equivalence, is that it preserves the above interpretation of nested expressions. We show next that a similar behavior can be obtained using standard 3-valued logic ($\L3$). Surprisingly, $\L3$ provides the same interpretation for nested expressions, but generally differs once free nesting of rule conditionals is allowed.

%%%%%%%%%%%%%%%%%%%%%%%%%%%%%%%%%%%%%%%%%%%%%%%%%%%%%%%%%%%%%%%%%%%%%%%%%%%%%%%
\section{$\L3$: Three valued logic.}
%%%%%%%%%%%%%%%%%%%%%%%%%%%%%%%%%%%%%%%%%%%%%%%%%%%%%%%%%%%%%%%%%%%%%%%%%%%%%%%
\label{sec:3v}

We will use propositional syntax plus Lukasiewicz's unary operator\footnote{For instance, see~\cite{Bul84}, pag. 8, where $\l$ is denoted as $\Box$.} $\l$. Intuitively, a formula $\l F$ is never unknown and points out that $F$ is valuated to true. In this way, $\neg \l F$ would mean that ``$F$ is not true,'' i.e., it is either false or unknown. If $F, G$ are $\L3$ formulas and $p$ an atom of the signature $\S$ then:
\begin{eqnarray*}
p, \ \neg F, \ F \vee G, \ \top, \ \bot, \ \l F
\end{eqnarray*}
\noindent are also $\L3$ formulas. Propositional derived operators ($\wedge, \supset, \equiv$) are defined in the usual way.

A three valued interpretation $M$ is a function $M:\S \longrightarrow \{0,\u,1\}$ assigning to each atom $p \in \S$ a truth value $M(p)$ which can be $0$ (false), $\u$ (unknown) or $1$ (true). We will usually represent $M$ as the pair of sets of atoms $(I^p,I^a)$ respectively containing the {\em positive} (true) and {\em consistent} (non-false) atoms where, of course, we require consistence: $I^p \subseteq I^a$. Consequently:
\begin{itemize}
\item[] $M(p)= \left\{ 
                \begin{array}{ll}
                1   & \ {\rm if} \ p \in I^p\\
                0   & \ {\rm if} \ p \not\in I^a\\
		\u  & \ {\rm otherwise}
                \end{array}
        \right.$
\end{itemize}

Note that we use here the same notation as for the pairs we handled in the reified approach. This is not casual: the negative information of a 3-valued interpretation will be used to represent default negation, whereas the positive information will represent the set of proved atoms.

\begin{df}{\bf ($\L3$ valuation of a formula)} \\
\label{df:3v}
We extend the valuation function $M$ to any formula $F$, $M(F) \in \{0,\u,1\}$, so that:
{\em
\begin{enumerate}
\item[1)] $M(\true) = 1$ and $M(\false) = 0$
\item[2)] $M(\neg F) = 1-M(F)$
\item[3)] $M(F \vee G) = max(M(F), M(G))$
\item[4)] $M(\l F)= \left\{ 
                \begin{array}{ll}
                1  & \ {\rm if} \ M(F)=1 \\
		0  & \ {\rm otherwise}
                \end{array}
        \right.$
\end{enumerate}
}
\end{df}

An interpretation $M$ {\em satisfies} a formula $F$, written $M \models_3 F$ when $M(F)=1$. When $F$ is satisfied by {\em any} interpretation, we call it an $\L3$-{\em tautology} and write $\models_3 F$. As usual, an interpretation is a {\em model} of a theory when it satisfies all its formulas. Maintaining the previous terminology, a 3-valued interpretation $M$ is called {\em total} iff it has the shape $M=(I,I)$, that is, it contains no unknown atoms. Clearly, when considering total interpretations, the $\l$ operator can be simply removed, and $\L3$ collapses into 2-valued propositional logic.

LP connectives are simply defined among the following derived operators:
\begin{eqnarray*}
\m F & \eqdef & \neg \l \neg F \\
\Not F & \eqdef & \neg \m F \\
G \leftarrow F & \eqdef & (\l F \supset \l G) \wedge 
                                   (\m F \supset \m G) \\
F \eq G & \eqdef & (F \leftarrow G) \wedge (G \leftarrow F)
\end{eqnarray*}
It is easy to check that the derived semantics for each one of these operators corresponds to:
\begin{itemize}
\item[5)] $M(\m F)= \left\{ 
                \begin{array}{ll}
                1  & \ {\rm if} \ M(F) \neq 0 \\
		0  & \ {\rm otherwise}
                \end{array}
        \right.$
\item[6)] $M(\Not F)= \left\{ 
                \begin{array}{ll}
                1  & \ {\rm if} \ M(F) = 0 \\
		0  & \ {\rm otherwise}
                \end{array}
        \right.$
\item[7)] $M(G \leftarrow F)= \left\{ 
                \begin{array}{ll}
                1  & \ {\rm if} \ M(F) \leq M(G) \\
		0  & \ {\rm otherwise}
                \end{array}
        \right.$
\item[8)] $M(F \eq G)= \left\{ 
                \begin{array}{ll}
                1  & \ {\rm if} \ M(F) = M(G) \\
		0  & \ {\rm otherwise}
                \end{array}
        \right.$
\end{itemize}

Operator $\m$ acts is the dual of $\l$, ($\m F$ can be read as ``$F$ is consistent") whereas implication $\leftarrow$ is the one proposed by Fitting~\cite{Fit85} and Kunen~\cite{Kun87}. When we represent some program $\Pi$ inside $\L3$, we will consider it as a single formula consisting in the conjunction of all the program rules. Note also that when $\models_3 F \eq G$, we can apply uniform substitution in $\L3$ as we would do in classical propositional logic. For instance, $\models_3 (\Not F) \eq (\bot \leftarrow F)$ means that we can replace any occurrence of $(\Not F)$ by $(\bot \leftarrow F)$ and vice versa. Let $\x$ and $\y$ be two meta-operators, any of them indistinctly representing $\l$ or $\m$. Then, the following formulas are also $\L3$ tautologies:
\begin{eqnarray}
\x (F \wedge G) & \eq & \x F \wedge \x G \label{f:p1}\\
\x (F \vee G) & \eq & \x F \vee \x G  \label{f:p2}\\
\y \x F & \eq & \x F \label{f:p3}\\
\y \neg \x F & \eq & \neg \x F \label{f:p4}
\end{eqnarray}

As $\m$ is defined in terms\footnote{Of course, we could have equally chosen the dual operator $\m$ as the basic one.} of $\l$, this means that we can unfold any $\L3$ formula until $\l$ is exclusively applied to literals. Using these properties, the following lemma can be easily proved:
\begin{lem}
\label{lem:L3interp}
Let $R$ be a pogram rule like (\ref{f:lprule}), and let the pair $M=(I^p,I^a)$ have the common shape of an $\L3$-interpretation and a classical interpretation for $proved$/$assumed$. Then, $M \models_3 R$ iff $M \models \ta{R} \wedge \tb{R}$.
\end{lem}

Besides, by inspection on $\L3$ semantics, we also have that:

\begin{lem}
\label{lem:unnest}
For any transformation $\alpha \EQ \beta$ in {\em (i)-(xii):} $\ \ \models_3 \alpha \eq \beta$.
\end{lem}

\begin{Th}
\label{th:strong3}
Let $\Pi_1$ and $\Pi_2$ be two logic programs possibly containing nested expressions. Then $\Pi_1$ and $\Pi_2$ are strongly equivalent iff: $\ \ \models_3 \l \Pi_1 \equiv \l \Pi_2$.
\end{Th}

Notice that we check $\l \Pi_1 \equiv \l \Pi_2$ instead of the stronger condition $\Pi_1 \eq \Pi_2$. To understand the difference, consider $\Pi_1=\{a\}$ and $\Pi_2=\{a \leftarrow \top\}$. The interpretation $M=(\{a\},\{a\})$ is the only model of both programs and so, $\models_3 \l \Pi_1 \equiv \l \Pi_2$. However, $\Pi_1 \eq \Pi_2$ is not a tautology, since $M'=(\emptyset,\{a\})$ makes $M'(\Pi_1) = \u \neq 1 = M'(\Pi_2)$.

%%%%%%%%%%%%%%%%%%%%%%%%%%%%%%%%%%%%%%%%%%%%%%%%%%%%%%%%%%%%%%%%%%%%%%%%%%%%%%%
\section{Differences with respect to HT}
%%%%%%%%%%%%%%%%%%%%%%%%%%%%%%%%%%%%%%%%%%%%%%%%%%%%%%%%%%%%%%%%%%%%%%%%%%%%%%%

Theorem \ref{th:strong3} shows that HT and $\L3$ coincide in their interpretations of programs with nested expressions. The next natural question is, do the HT and $\L3$ interpretations coincide for any arbitrary theory? The answer to this question is negative, as we will show with a pair of counterexamples. Of course, due to theorem \ref{th:strong3}, these counterexamples cannot be just programs with nested expressions, as defined in Section \ref{sec:nest}. We study, for instance, a nested conditional, and the negation of a conditional.

Consider the theory consisting of the singleton formula $(a \leftarrow b) \leftarrow c$. In HT, this theory is equivalent to $(a \leftarrow b, c)$, which seems to be the most intuitive solution, whereas in $\L3$ it is actually equivalent to $(a; \Not c \leftarrow b)$. Both equivalences hold in classical propositional logic. However, for computing stable models, their behavior is quite different. For instance, the theory $\{b, (a \leftarrow c), (c \leftarrow a), ((a \leftarrow b) \leftarrow c)\}$ would have a unique stable model $\{b\}$ under the HT interpretation whereas, under $\L3$, an additional stable model $\{a,b,c\}$ is obtained.

The second example shows the most important problem of the $\L3$ interpretation: once we allow arbitrary theories, we may obtain non-subtotal models, something that does not happen\footnote{See for instance Fact 1 in~\cite{Lif00}.} in HT. Let $\Pi$ be the theory $\{b, \Not (a \leftarrow b)\}$. Its unique stable model is $\{b\}$ both in HT and $\L3$. However, while the pair $(\{b\},\{b\})$ is the unique HT model\footnote{In fact, the expression $\Not (a \leftarrow b)$ is HT-equivalent to the pair of constraints $(\bot \leftarrow \Not b)$ and $(\bot \leftarrow a)$.} of $\Pi$, in $\L3$ there exists a second model $(\{b\},\{a,b\})$ which is not subtotal. In other words, when using $\L3$ for this general syntax, the set of $\L3$ models does not fully characterize strong equivalence.

%%%%%%%%%%%%%%%%%%%%%%%%%%%%%%%%%%%%%%%%%%%%%%%%%%%%%%%%%%%%%%%%%%%%%%%%%%%%%%%
\section{Discussion}
%%%%%%%%%%%%%%%%%%%%%%%%%%%%%%%%%%%%%%%%%%%%%%%%%%%%%%%%%%%%%%%%%%%%%%%%%%%%%%%
\label{sec:conc}

The study of strong equivalence is probably one of the most active current topics in research in Logic Programming, as it becomes evident by the increasing amount of new results obtained recently (just to cite three examples~\cite{Tur01,Pea01,Jon01}). 

In~\cite{Pea01}, a classical logic characterization is also provided, which presents several similarities with the approach we present here. The main difference of Pearce et al's method is that it actually relies on a syntactic translation from HT into classical logic. This translation informally consists in a duplication of the atoms in the signature so that an atom $p$ denotes our $proved$ whereas an atom $p'$ would denote $assumed$. In this paper, our initial motivation for using classical logic was to improve the presentation and the understanding. In this way, we have directly started from non-nested programs, trying to capture the definition of stable models in a way as direct as possible. As a result, our characterization does not provide an interpretation of nested connectives. In order to deal with them, we would need to apply a previous step, using transformations (i)-(xii). Pearce et al's encoding starts from HT logic, and so, deals with nested expressions (in the same way as HT does). Besides, the transformation presented in~\cite{Pea01} has the additional advantage of being linear, while (i)-(xii) are not polynomial in the general case. Despite of these two advantages of Pearce et al's approach, it must be noticed that none of the two classical encodings can actually be considered a full-semantics for nested logic programs, since {\em in both cases}, a previous syntactic transformation is required. Therefore, translation to classical logic is very interesting for practical purposes, but is limited from a purely semantic point of view.

Another similarity between our classical encoding with respect to~\cite{Pea01} is, not only how to decide strong equivalence, but how to obtain stable models. In our Section~\ref{sec:class} we simply used to that purpose the result presented by Lin and Shoham in~\cite{Lin92} and then included slight variations that we proved to be sound. In~\cite{Pea01}, a quantified boolean formula is used instead:
\begin{eqnarray}
\phi' \wedge \neg \exists V ((V < V') \wedge \tau_{HT}[\phi]) \label{f:QBF}
\end{eqnarray}
\noindent where $V$ is the set of atoms, $\phi$ is the original program, $\phi'$ results from replacing any atom $p$ by $p'$ and finally $\tau_{HT}[\phi]$ is Pearce et al's translation from HT to classical logic. On the other hand, Lin and Shoham's result involving circumscription can be formulated\footnote{As described for instance in~\cite{Lif93}, propositional circumscription is nothing else but a quantified boolean formula.} as:
\begin{eqnarray}
(V=V') \wedge {\cal C}[\phi] \wedge \neg \exists V ((V < V') \wedge {\cal C}[\phi]) \label{f:circ}
\end{eqnarray}
\noindent where ${\cal C}[\phi]$ simply replaces each $\Not p$ by $\neg p'$. Notice how, at least structurally, (\ref{f:circ}) is very similar to (\ref{f:QBF}).

As for the $\L3$ encoding, it must also be noticed that other logical characterizations have been obtained apart from HT. In~\cite{Jon01}, for instance, they use instead another logic, KC, and show that this is, in fact, the weakest intermediate logic (between intuitionistic and classical) that allows capturing strong equivalence of logic programs with nested expressions. An interesting open question is how logic KC deals with nested conditionals since, as we have shown, this is the case where HT and $\L3$ diverge.

The interest of the encoding we have presented is perhaps the use of a multi-valued logic which is commonly known and understood and yields a natural characterization. We believe that the use of $\L3$ may be especially suitable as a valid alternative for ``by hand" studies of strong equivalence, since it allows applying intuitive properties of $\l$ and $\m$ operators (as an example, look at the proofs in the appendix) in a more or less intuitive way.

\subsubsection*{Acknowledgements} 
I want to thank Vladimir Lifschitz for his discussions and comments about a preliminary draft of the $\L3$ encoding, and to the anonymous referees for drawing my attention to part of the related work cited in this paper. This research is partially supported by the Government of Spain, grant {TIC2001-0393}.
\bibliographystyle{plain}
\bibliography{sm3v}

%%%%%%%%%%%%%%%%%%%%%%%%%%%%%%%%%%%%%%%%%%%%%%%%%%%%%%%%%%%%%%%%%%%%%%%%%%%%%%%
\APPEND{A}{Proofs}
%%%%%%%%%%%%%%%%%%%%%%%%%%%%%%%%%%%%%%%%%%%%%%%%%%%%%%%%%%%%%%%%%%%%%%%%%%%%%%%

\noindent {\bf Proof of theorem \ref{th:strong}}\\
The right to left direction was proved in proposition \ref{ppn:suff}. For the left to right direction, assume there is some $M=(I^p,I^a)$, $M \in \SUBT(\ta{\Pi}_1)$ but $M \not\in \SUBT(\ta{\Pi}_2)$. Let $M'$ be the total pair $M'=(I^a,I^a)$. We will show that it is possible to construct a program $\Pi$ so that $I^a$ is a stable model of $\Pi_1 \cup \Pi$ but not of $\Pi_2 \cup \Pi$, or vice versa. We consider two cases:
\begin{enumerate}
\item $M' \not\in \SUBT(\ta{\Pi}_2)$. As $M \in \SUBT(\ta{\Pi}_1)$, by definition of $\SUBT$, $M'$ is model of $\ta{\Pi}_1$. We construct the program $\Pi = \{p : p \in I^a\}$. Clearly, $M'$ is still model of $\ta{\Pi_1} \wedge \ta{\Pi}$ but, now, it is also minimal: a strict subset of $proved$ atoms is not allowed. Therefore, $I^a$ is a stable model of $\Pi_1 \cup \Pi$. On the other hand, as $M'$ is total, but $M' \not\in \SUBT(\ta{\Pi}_2)$, then $M' \not\models \ta{\Pi}_2$. As a result, $I^a$ cannot be stable model of $\Pi_2 \cup \Pi$.

\item $M' \in \SUBT(\ta{\Pi}_2)$. Note first that, as $M \not\in \SUBT(\ta{\Pi}_2)$ and $M \preceq M'$ we conclude $M \not\models \ta{\Pi}_2$ and $M \prec M'$ . We construct the program:
\begin{eqnarray*}
\Pi & = & \{p : p \in I^p\} \cup \\
    &   & \{(p \leftarrow q) : p,q \in (I^a - I^p), p \neq q\}
\end{eqnarray*}
\noindent It is easy to see that $M' \models \ta{\Pi}$ and so, $M' \models \ta{\Pi}_2 \wedge \ta{\Pi}$. We will show that $M'$ is also minimal for that theory. Let us assume that there exists another model $(J,I^a)$ strictly lower than $M'$, $J \subset I^a$, and try to reach a contradiction. As atoms of $I^p$ are facts of $\Pi$, but $M=(I^p,I^a)$ is not model of $\ta{\Pi}_2$, we must have: $I^p \subset J$. Now, take some $q \in J-I^p$ and some $p \in I^a-J$. The rule $p \leftarrow q$ belongs to $\Pi$ but $(J,I^a)$ does not satisfy the rule translation. Therefore, $(J,I^a) \not\models \ta{\Pi}$ and we reach a contradiction. Since we have proved that $M'$ is minimal, $I^a$ is stable model of $\ta{\Pi}_2 \cup \ta{\Pi}$. Finally, we study program $\Pi_1$. It can be easily checked that $M \models \ta{\Pi}$ and so $M \models \ta{\Pi}_1 \cup \ta{\Pi}$. But as $M \prec M'$, $M'$ is not minimal for $\ta{\Pi}_1 \cup \ta{\Pi}$ and so $I^a$ is not stable model for this program.
\end{enumerate}

\vspace{20pt}
\noindent {\bf Proof of theorem \ref{th:circ}}\\
We start from $\CIRC[\Pi^*;proved] \wedge (\ref{f:circ-2})$, observing that the conjuncts $\tb{\Pi}$ and (\ref{f:circ-4}) of $\Pi^*$ can be moved outside the circumscription. The reason for this is that in $\tb{\Pi}$, predicate $proved$ does not occur, whereas in (\ref{f:circ-4}) it occurs negatively (considering $\alpha \supset \beta$ as $\neg \alpha \vee \beta$). Therefore, we can respectively apply equivalence (3.2) and proposition 3.3.1 in~\cite{Lif93}, to obtain the equivalent formula:
\begin{eqnarray*}
\CIRC[\ta{\Pi} ;proved] \wedge \tb{\Pi} \wedge (\ref{f:circ-4}) \wedge (\ref{f:circ-2})
\end{eqnarray*}
Now, notice that, by definition, the circumscription of $\CIRC[\ta{\Pi};proved]$ has the shape of a conjunction $\ta{\Pi} \wedge \alpha$ (where $\alpha$ is a second order formula we do not need to detail). So, the above formula is actually:
\begin{eqnarray*}
\ta{\Pi} \wedge \alpha \wedge \tb{\Pi} \wedge (\ref{f:circ-4}) \wedge (\ref{f:circ-2})
\end{eqnarray*}
Finally, notice that (\ref{f:circ-2}) implies (\ref{f:circ-4}), whereas $\ta{\Pi} \wedge (\ref{f:circ-2})$ implies $\tb{\Pi}$, and so, the implied conjuncts can be removed, obtaining the equivalent formula:
\begin{eqnarray*}
\ta{\Pi} \wedge \alpha \wedge (\ref{f:circ-2})
\end{eqnarray*}
\noindent which is nothing else but $\CIRC[\ta{\Pi};proved] \wedge (\ref{f:circ-2})$.

\vspace{20pt}
\noindent {\bf Proof of property \ref{ppy:total}}\\
We must show that $M'=(I^a,I^a)$ is also model of $\Pi^*$. Clearly, as $\tb{\Pi}$ exclusively refers to predicate $assumed$, $M \models \tb{\Pi}$ iff $M' \models \tb{\Pi}$. On the other hand, it is clear that $M' \models (\ref{f:circ-4})$ -- in fact, it further satisfies (\ref{f:circ-2}). So, we only have to prove that $M' \models \ta{\Pi}$. Assume that, for some rule $R \in \Pi$, $M' \not\models \ta{R}$, which in classical logic is equivalent to $M' \models \neg (\ta{R})$, i.e., $M'$ must satisfy:
\begin{multline*}
\PAR{\bigwedge^r_{i=1} proved(c_i)} \wedge \PAR{\bigwedge^s_{i=1} \neg assumed(d_i)} \wedge \\  \PAR{\bigwedge^n_{i=1} \neg proved(a_i) } \wedge \PAR{\bigwedge^m_{i=1} assumed(b_i)} 
\end{multline*}
As in $M'$, $proved$ and $assumed$ are equivalent, this implies that $M'$ must also satisfy:
\begin{multline*}
\PAR{\bigwedge^r_{i=1} assumed(c_i)} \wedge \PAR{\bigwedge^s_{i=1} \neg assumed(d_i)} \wedge \\  \PAR{\bigwedge^n_{i=1} \neg assumed(a_i) } \wedge \PAR{\bigwedge^m_{i=1} assumed(b_i)}
\end{multline*}
But this formula is exactly $\neg (\tb{R})$ and so, $M'$ cannot be model of $\tb{\Pi}$ nor $\Pi^*$, reaching a contradiction.

\vspace{20pt}
\noindent {\bf Proof of lemma \ref{lem:L3interp}}\\
Using properties (\ref{f:p1})-(\ref{f:p4}) and the definitions of derived operators, a general logic program rule like (\ref{f:lprule}) is equivalent to the conjunction of the pair of formulas:
\begin{multline}
\l c_1 \wedge \dots \wedge \l c_r \wedge \neg \m d_1 \wedge \dots \wedge \neg \m d_s \supset \\
\l a_1 \vee \dots \vee \l a_n \vee \dots \vee \neg \m b_1 \vee \dots \vee \neg \m b_m \label{f:l3rulea}
\end{multline}
\begin{multline}
\m c_1 \wedge \dots \wedge \m c_r \wedge \neg \m d_1 \wedge \dots \wedge \neg \m d_s \supset \\
\m a_1 \vee \dots \vee \m a_n \vee \dots \vee \neg \m b_1 \vee \dots \vee \neg \m b_m \label{f:l3ruleb}
\end{multline}
Note that in formulas, operators $\l$ and $\m$ are only applied to atoms and, vice versa, atoms only appear in the scope of $\l$ and $\m$. As the interpretation of $\l$ and $\m$ is always bivalued, we can just consider the meaning of each $\l p$ and $\m p$ individually and then interprete the whole formula as a classical propositional one. Now, note that $\l p$ is true iff $p \not\in I^p$ that is $M \models proved(a)$, and $\m p$ is true iff $p \in I^a$, that is, $M \models assumed(a)$. Therefore, (\ref{f:l3rulea}) and (\ref{f:l3ruleb}) respectively correspond to $\ta{R}$ and $\tb{R}$.

\vspace{20pt}
\noindent {\bf Proof of lemma \ref{lem:unnest}}\\
As `,' and `;' are simply understood as $\wedge$ and $\vee$, properties (i)-(iii), (vi) and (vii) are straightforward.
\begin{enumerate}
\item[(iv)] The formula $\Not (F; G)$ is, by definition, $\neg \m (F \vee G)$ which, by (\ref{f:p2}), is always valuated in the same way as $\neg (\m F \vee \m G)$. As $\L3$ satisfies the De Morgan's laws, this is equivalent to $\neg \m F \wedge \neg \m G$, which by definition, is the formula $\Not F , \Not G$. The other part of property (iv) is analogous.
\item[(v)] The formula $\Not \Not \Not F$ is, by definition, $\neg \m \neg \m \neg \m F$. By (\ref{f:p4}), we replace $\m \neg \m F$ by $\neg \m F$ obtaining $\neg \m \neg \neg \m F$, that is, $\neg \m \m F$. Now, by (\ref{f:p3}) this is equivalent to $\neg \m F$, which is, by definition $\Not F$.
\item[(viii)] The formula $\Not \bot$ is $\neg \m \bot$. As $\m \bot$ is always valuated as 0 (false), this formula is equivalent to $\neg \bot$, that is, $\top$. The other part of (viii) is analogous.
\item[(ix)] The formula $(F,G \leftarrow H)$ is, by definition:
\begin{eqnarray*}
(\l H \supset \l (F \wedge G)) \wedge (\m H \supset \m (F \wedge G))
\end{eqnarray*}
Applying (\ref{f:p1}) we obtain:
\begin{eqnarray*}
(\l H \supset \l F \wedge \l G) \wedge (\m H \supset \m F \wedge \m G)
\end{eqnarray*}
This is equivalent to:
\begin{multline*}
(\neg \l H \vee \l F) \wedge (\neg \l H \vee \l G) \wedge \\
(\neg \m H \vee \m F) \wedge (\neg \m H \vee \m G)
\end{multline*}
\noindent (simply unfold the definition of $\supset$ and apply distributivity). Finally, note that, by definition of $\leftarrow$, the last formula is just:
\begin{eqnarray*}
(F \leftarrow H) \wedge (G \leftarrow H)
\end{eqnarray*}
\item[(ix)] We just provide a proof sketch:

\begin{tabular}{ll}
$(\l (G \vee H) \supset \l F) \wedge$ \\
$\ \ (\m (G \vee H) \supset \m F)$ & Def. of $\leftarrow$ \\
\\
$(\l G \vee \l H \supset \l F) \wedge$ \\
$\ \ (\m G \vee \m H \supset \m F)$ & (\ref{f:p2}) \\
\\
$(\l G \supset \l F) \wedge (\m G \supset \m F) \wedge$ & \\
$\ \ (\l H \supset \l F) \wedge (\m H \supset \m F)$ & Def. of $\supset$ \\
\\
$(F \leftarrow G) \wedge (F \leftarrow H)$ & Def. of $\leftarrow$
\end{tabular}

\item[(x)] Again, as a proof sketch:

\begin{tabular}{ll}
$(\l (G \wedge \neg \m \neg \m H) \supset \l F) \wedge $ \\
$\ \ (\m (G \wedge \neg \m \neg \m H) \supset \m F)$ & Def. of $\leftarrow$, $\Not$ \\
$(\l (G \wedge \neg \neg \m H) \supset \l F) \wedge$ \\
$\ \ (\m (G \wedge \neg \neg \m H) \supset \m F)$ & (\ref{f:p4}) \\
$(\l G \wedge \l \m H \supset \l F) \wedge$ \\
$\ \ (\m G \wedge \m \m H \supset \m F)$ & (\ref{f:p1}) \\
$(\l G \wedge \m H \supset \l F) \wedge $ \\
$\ \ (\m G \wedge \m H \supset \m F)$ & (\ref{f:p3}) \\
$(\l G \supset \l F \vee \neg \m H) \wedge $\\
$\ \ (\m G \supset \m F \vee \neg \m H)$ & Def. $\supset$ \\
$F; \Not H \leftarrow G$ & Def. of $\leftarrow$, $\Not$
\end{tabular}

\item[(xi)] It is analogous to proof for (x).
\end{enumerate}

\vspace{20pt}
\noindent {\bf Proof of theorem \ref{th:strong3}}\\
First, from lemma \ref{lem:unnest} the corresponding non-nested program has the same set of $\L3$ models. Second, lemma \ref{lem:L3interp} shows that this program has the same models than our classical encoding. Therefore, we can directly apply theorem \ref{th:strong2}.

%%%%%%%%%%%%%%%%%%%%%%%%%%%%%%%%%%%%%%%%%%%%%%%%%%%%%%%%%%%%%%%%%%%%%%%%%%%%%%%
\end{document}